\pdfoutput=1

\documentclass[11pt]{article}

\usepackage{acl}

\usepackage{times}
\usepackage{latexsym, booktabs}

\usepackage[T1]{fontenc}

\usepackage[utf8]{inputenc}

\usepackage{microtype}

\usepackage{todonotes}
\usepackage{multirow, subcaption}
\usepackage[noend,ruled]{algorithm2e}
\title{Improving End-to-End Speech Translation by Imitation-Based Knowledge Distillation with Synthetic Transcripts} 

\author{Rebekka Hubert\thanks{\hspace{1ex}All work was done at Heidelberg University.} \\
  Computational Linguistics\\
  Heidelberg University, Germany \\
  \texttt{\small hubert@cl.uni-heidelberg.de} \\\And
  Artem Sokolov \\
  Google Research \\
  Berlin, Germany \\
  \texttt{\small artemsok@google.com} \\ \And
  Stefan Riezler \\
  Computational Linguistics \& IWR\\
  Heidelberg University, Germany \\
  \texttt{\small riezler@cl.uni-heidelberg.de}
  }

\usepackage{algorithm2e}      
\usepackage{amsmath}
\DeclareMathOperator*{\argmax}{argmax}
\usepackage{subcaption}
\usepackage{graphicx, xcolor, xspace}
\usepackage{csquotes}
\usepackage{amsfonts, verbatim}
\usepackage[nolist, nohyperlinks]{acronym}
\usepackage{todonotes}

\begin{document}
\begin{acronym}
\acro{il}[IL]{Imitation Learning}
\acro{imitkd}[ImitKD]{Imitation-based Knowledge Distillation}
\acro{asrimitkd}[SynthImitKD]{Imitation-based Knowledge Distillation with synthetic transcripts}
\acro{kd}[KD]{Knowledge Distillation}
\acro{asrkd}[SynthKD]{Knowledge Distillation with synthetic transcripts}
\acro{rnn}[RNN]{Recurrent Neural Network}
\acro{skd}[SKD]{sequence-level knowledge distillation}
\acro{wer}[WER]{Word Error Rate}
\acro{wkd}[WKD]{word-level knowledge distillation}
\end{acronym}

\maketitle

\begin{abstract}

End-to-end automatic speech translation (AST) relies on data that combines audio inputs with text translation outputs. 
Previous work used existing large parallel corpora of transcriptions and translations in a knowledge distillation (KD) setup to distill a neural machine translation (NMT) into an AST student model. 
While KD allows using larger pretrained models, the reliance of previous KD approaches on manual audio transcripts in the data pipeline restricts the applicability of this framework to AST. We present an imitation learning approach where a teacher NMT system corrects the errors of an AST student without relying on manual transcripts. We show that the NMT teacher can recover from errors in automatic transcriptions and is able to correct erroneous translations of the AST student, leading to improvements of about 4 BLEU points over the standard AST end-to-end baseline on the English-German CoVoST-2 and MuST-C datasets, respectively. Code and data are publicly available.\footnote{\url{https://github.com/HubReb/imitkd_ast/releases/tag/v1.1}}
\end{abstract}

\section{Introduction}


The success of data-hungry end-to-end automatic speech translation (AST) depends on large amounts of data that consist of speech inputs and corresponding translations. One way to overcome the data scarcity issue is a knowledge distillation (KD) setup where a neural machine translation (NMT) expert (also called oracle) is distilled into an AST student model~\citep{liu19d_interspeech,gaido-etal-2020-end}. 
The focus of our work is the question of whether the requirement of high-quality source language transcripts, as in previous applications of KD to AST, can be relaxed in order to enable a wider applicability of this setup to AST scenarios where no manual source transcripts are available. 
Examples for such scenarios are low-resource settings (e.g., for languages without written form for which mostly only audio-translation data are available), or settings where one of the main uses of source transcripts in AST --- pre-training the AST encoder from an automatic speech recognition (ASR) system--- is replaced by a large-scale pre-trained ASR system (which itself is trained on hundreds of thousands hours of speech, but the original training transcripts are not available \citep{RadfordETAL:22,ZhangETAL:22}). Relaxing the dependence of pre-training AST encoders on manual transcripts has recently been studied by \citet{zhang2022revisiting}. 
Our focus is instead to investigate the influence of manual versus synthetic transcripts as input to the teacher model in an imitation learning (IL) approach \citep{lin-etal-2020-autoregressive, DBLP:journals/corr/abs-2109-04114}, and  to lift this scenario to AST.  To our knowledge, this has not been attempted before. We present a proof-of-concept experiment where we train an ASR model on a few hundred hours of speech, but discard the manual transcripts in IL training, and show that this ASR model is sufficient to enable large NMT models 
to function as error-correcting oracle in an IL setup. 
Focusing on the IL scenario, we show that one of the key ingredients to make our framework perform on synthetic ASR transcripts is to give the AST student access to the oracle's full probability distribution instead of only the expert's optimal actions.
Furthermore, when comparing two IL algorithms of different power --- either correcting the student output in a single step, or repairing outputs till the end of the sequence --- we find that, at least in the setup of a reference-agnostic NMT teacher, the single-step correction of student errors is sufficient. 

One of the general reasons for the success of our setup may be a reduction of data complexity and an increase of variations of outputs, similar to applications of KD in NMT \citep{ZhouETAL:20}. 
To investigate the special case of imitation-based KD on synthetic speech inputs, we provide a manual analysis of the NMT expert's behavior when faced with incorrect synthetic transcripts as input, or when having to correct a weak student's translation in the IL setting. We find that the NMT oracle can correct errors even if the source language input lacks semantically correct information, by utilizing its language modeling capability to correct the next-step token. This points to new uses of large pre-trained ASR and NMT models (besides initialization of encoder and decoder, respectively) as tools to improve non-cascading end-to-end AST.

\section{Related Work} 

Imitation learning addresses a deficiency of sequence-to-sequence learning approaches, nicknamed \textit{exposure bias}~\citep{NIPS2015_e995f98d,Ranzato2016SequenceLT}, that manifests as the inference-time inability to recover from own errors, leading to disfluent or hallucinated translations~\citep{sennrich}. 
IL aims to replace the standard learning paradigm of teacher forcing~\cite{oldrnnpaper} (which decomposes sequence learning into independent per-step predictions, each conditioned on the golden truth context rather than the context the model would have produced on its own) by enriching the training data with examples of successful recovery from errors. We build upon two previous adaptations of IL to NMT~\cite{lin-etal-2020-autoregressive, DBLP:journals/corr/abs-2109-04114} and lift them to AST. 

Knowledge distillation \citep{hinton2015distilling} transfers the knowledge encoded in a large model, called teacher, to a far smaller student model by using the teacher to create soft labels and train the student model to minimize the cross-entropy to the teacher. KD has been successfully used for machine translation \citep{kim-rush-2016-sequence}, speech recognition \citep{wong16_interspeech} and speech translation \citep{liu19d_interspeech}.

Synthetic speech translation training datasets have been used previously to train AST models:
\citet{pino20_interspeech} used an ASR-NMT model cascade to translate unlabeled speech data for augmentation. To obtain more machine translation (MT) training data,~\citet{jia2019leveraging, pino-etal-2019-harnessing} generated synthetic speech data with a text-to-speech model. 
\citet{liu19d_interspeech} applied KD between an NMT expert and an AST student with manual transcriptions as expert input to improve AST performance. 
\citet{gaido-etal-2020-end} improved upon this by increasing the available training data by utilizing a MT model to translate the audio transcripts of ASR datasets into another language, yet they still use manual transcripts for distillation in the following finetuning phase.

Further attempts focused on improving AST models by utilizing MT data for multitask learning with speech and text data~\citep{Tang2021AGM, tang-etal-2021-improving,Bahar2019ACS, DBLP:conf/interspeech/WeissCJWC17, anastasopoulos-chiang-2018-tied}, such as XSTNet~\citep{ye2021end} and FAT-MLM~\citep{zheng2021fused}.

A question orthogonal to ours, concerning the influence of pre-training encoder and/or decoder on source transcripts, has been investigated by \citet{zhang2022revisiting}. 
They achieved  competitive results without any pretraining via the introduction of parameterized distance penalty and neural acoustic feature modeling in combination with CTC regularization with translations as labels. Their question and solutions are orthogonal to ours and are likely to be yield independent benefits. 
 


\section{Imitation-based Knowledge Distillation}
\label{sec:ikd}

We view an auto-regressive NMT or AST system as a policy $\pi$ that defines a conditional distribution over a vocabulary of target tokens $v\in V$ that is conditioned on the input $x$ and the so far generated prefix $y_{<t}$: $\pi(v|y_{<t}; x)$. This policy is instantiated as the output of the softmax layer. 
When training with teacher-forcing, the cross-entropy (CE) loss  $\ell(\cdot)$ is minimized under the \emph{empirical} distribution of training data $D$: $\mathcal{L_{\text{CE}}(\pi)} = \mathbb{E}_{(y, x)\sim D}[\sum_{t=1}^T \ell(y_t, \pi)]$. To perform well at test time we are interested in the expected loss under the \emph{learned} model distribution: $\mathcal{L(\pi)} = \mathbb{E}_{(y,x)\sim\pi}[\sum_{t=1}^T \ell(y_t, \pi)]$.

As shown by \citet{pmlr-v15-ross11a}, the discrepancy between $\mathcal{L}$ and $\mathcal{L_{\text{CE}}}$ accumulates quadratically with the sequence length $T$, which in practice could manifest itself as translation errors.
They proposed the Dagger algorithm which has linear worst-case error accumulation. It, however, relies on the existence of an oracle policy $\pi^\ast$ that, conditioned on the same input $x$ and the partially generated $\pi$'s prefix $y_{<t}$, can produce a single next-step correction to $y_{<t}$.
\citet{ross2} further proposed the AggreVaTe algorithm which relies on an even more powerful oracle that can produce a full continuation in the task-loss optimal fashion: For NMT, this means continuing the $y_{<t}$ in a way that maximizes BLEU, as done for example in~\citet{DBLP:journals/corr/abs-2109-04114}. 

\begin{figure}[t!]
		\centering
       \includegraphics[scale=0.3]{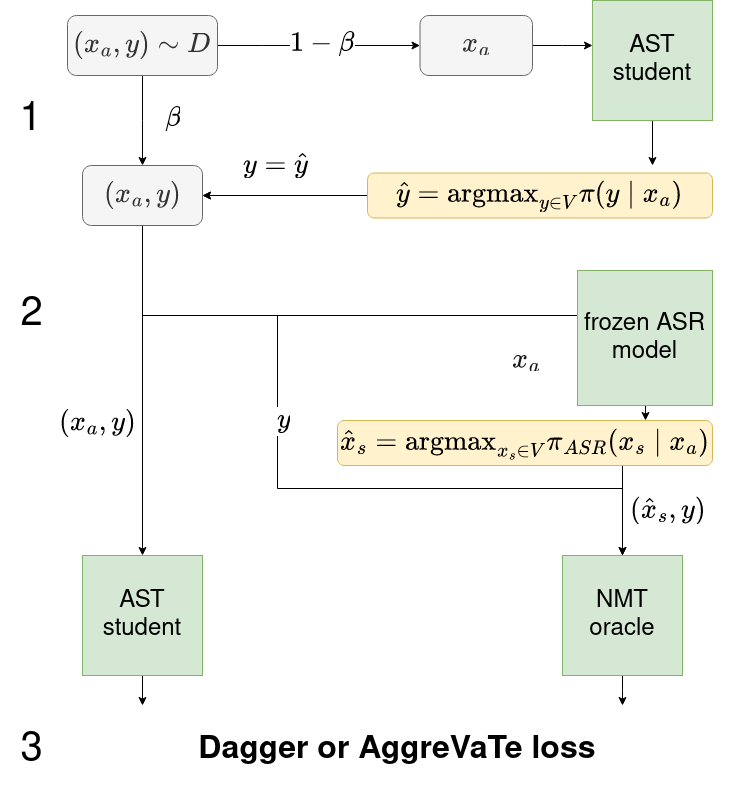}
    \caption{\textbf{Diagram of AST training with imitation learning and synthetic transcripts coming from ASR models}. (1) With probability $1 - \beta$ the AST student creates a hypothesis $\hat{y}$ that replaces the reference translation $y$. (2) The ASR model generates the synthetic transcript $\hat{x}_s$ for the audio sample $x_a$ to feed the NMT oracle as input. (3) Calculation of Dagger or AggreVaTe loss as shown in Algorithm \ref{alg:pseudo}.}
	\label{fig:asrimitkd}
\end{figure}

\paragraph{IL for NMT} 

We pretrain a large NMT model to serve as an oracle $\pi^\ast$ that either simply predicts the next-step optimal output vocabulary token $v^{\ast}_t$ given a source sentence $x$ and any (potentially, erroneous) partial student hypothesis $y_{<t}$ (Dagger):
\begin{equation}
v^{\ast}_t = \argmax\limits_{v\in V} \pi^\ast(v\mid y_{<t}; x),\label{eq:expert_task}
\end{equation}
or continues $y_{<t}$ till the end (AggreVaTe):
\begin{equation}
y^\ast_{> t} = \argmax_{y_{> t}} \pi^\ast(y_{<t} + a_t +y_{>t}\mid y_{<t}; x),\label{eq:AggreVaTe_oracle}
\end{equation}
where $y_{>t}$ is the continuation, $a_t$ is an exploratory action, and the last $\argmax$ is implemented as beam search.
The predicted $v^{\ast}_t$ or $y^\ast_{>t}$ are viewed as one-step or multi-step corrections of the current policy, and the student is updated to increase the probability of the correction via the cross-entropy loss on triples $(y_{<t}, x, v^{\ast}_t)$ in case of Dagger, or to decrease a square loss between logit $Q$ of the selected action $a_t$ and the BLEU of the predicted suffix\footnote{We use the difference between the BLEU values of the full sequence and that of the prefix~\cite{bahdanau16}.} from that action in case of AggreVaTe. 

Both algorithms proceed iteratively, where the newly generated set of triples form a provisional training data set $D_i$. Originally, Dagger and AggreVaTe train the student's $\pi_i$ on the aggregated dataset $\cup_{j\leq i} D_j$ and use a probabilistic mixture for the current roll-out policy, which queries the oracle with probability $\beta_i$ and the student otherwise. This setup guarantees that the prediction error scales at most linearly with time, unlike the quadratic scaling of the standard teacher forcing~\cite{pmlr-v15-ross11a}, which is standardly used in sequence-level KD. This makes Dagger and AggreVaTe  promising candidates to improve over KD. 

In our implementation, we follow~\citet{lin-etal-2020-autoregressive}, who save memory via training on individual $D_i$ in each iteration $i$, instead of training on the set union. They further speed up training by keeping the reference translation $y$ with probability $\beta_i$, and otherwise generate a translation $\hat{y}$ of the source sentence $x$ from the student policy (see Algorithm~\ref{alg:pseudo}). For each $t$ in the algorithm, AggreVaTe needs to generate an exploration token $a_t$ and calculate the BLEU it would lead to, according to the oracle continuation starting off this action. 

\begin{algorithm}[t!]
        \SetInd{0.45em}{0.45em}
        \small
	\KwData{Let $D$ be original bi-text dataset, $\pi^\ast$ the NMT oracle policy, $I$ the total number of iterations, $T$ the max sequence length, $Q$ the final logits, and $B$ the batch size.}
	Initialize $\pi_1$ arbitrarily.\\
	\For{$i = 1\dots I$}{
		Initialize $D_i \leftarrow \emptyset$\\
		\For{$b =1\dots B$}{
			Sample an example $(x, y) \sim D$.\\
			Sample uniformly $u \sim [0, 1]$\\
			\If{$u > \beta_i$}{
				Generate $\hat{y}$ from $\pi_i$ given $x$.\\
				Replace $y$ with $\hat{y}$.
			}
      \If{Dagger}{
			\For{$t=1\dots T$} {

    		    Predict $v^{\ast}_t = \argmax\limits_{v \in V} \pi^\ast(v \mid y_{<t}; x)$\\
                    Append $(y_{<t}, x, v^{\ast}_t)$ to $D_i$
  		    
               } 
        }
        \Else(\tcp*[h]{AggreVaTe}){
                Sample uniformly $t \in \lbrace 1, .., T \rbrace$.\\
               Predict $a_t = \argmax\limits_{v \in V} \pi(v \mid y_{<t}; x)$\\
               Predict $y^\ast_{>t} = \argmax\limits_{y_{>t}} \pi^\ast(y_{>t} \mid y_{<t}+a_t; x)$\\
               Append $(y_{<t}, x, a_t, \text{BLEU}(y^\ast_{>t}))$ to $D_i$}
		}
	    $\mathcal{L}_{\textrm{Dagger}}= \mathbb{E}_{D_i}\left[ - \sum\limits_{t=1}^T \log \pi_i (v^{\ast}_t\mid y_{<t}; x)\right]$\\
            $\mathcal{L}_{\textrm{AggreVaTe}}= \mathbb{E}_{D_i}\left[ \sum\limits_{t=1}^T \big(\sigma(Q(a_t\mid y_{<t};x))-\text{BLEU}(y^\ast_{>t})\big)^2\right]$\\
		Let $\pi_{i+1} = \pi_i - \alpha_i \cdot \frac{\partial \mathcal{L}}{\partial\pi_i}$.
	
 }
	\caption{Dagger/AggreVaTe for distillation in NMT; combined from \citep{lin-etal-2020-autoregressive} and \citep{DBLP:journals/corr/abs-2109-04114}.}
	\label{alg:pseudo}
\end{algorithm}

\paragraph{IL for AST} Adapting Dagger and AggreVaTe to an AST student is relatively straightforward (see Figure~\ref{fig:asrimitkd}): 
We feed the NMT oracle the source language transcript $x_s$ of the audio data sample $x_a$ that is also given to the AST student. We define an algorithm IKD (\text{imitation knowledge distillation}) that optimizes the cross-entropy of the student's policy w.r.t.~the optimal expert prediction:
\begin{equation}
	\mathcal{L}_{\text{IKD}}(\pi) = \mathbb{E}
 \left[ - \sum\limits_{t=1}^T  \log \pi (v^{\ast}_t \mid y_{<t}; x_a)\right],  \label{eq:imit_opt}
\end{equation}
with $v^{\ast}_t$ as in~\eqref{eq:expert_task}.
Algorithm IKD$^+$ optimizes the cross-entropy w.r.t.~the expert's policy:


\begin{align}
    & \mathcal{L}_{\text{IKD}^+}(\pi) = \\ \notag
    & \mathbb{E} \left[- \sum\limits_{v \in V} \pi^\ast (v \mid y_{< t}; x_s) \cdot \log \pi(v \mid y_{< t}; x_a) \right].
\end{align}

An important modification to these objectives that we propose in this work is to replace the gold source language transcripts $x_s$ fed to the NMT oracle by synthetic transcripts generated by a pretrained ASR model. We call this algorithm \mbox{SynthIKD}, with a respective SynthIKD$^+$ variant. 

\section{Experiments}\label{sec:experiments}

We experiment with English-German AST on the CoVoST2~\citep{wang21s_interspeech} (430 hours) and the MuST-C~\citep{di-gangi-etal-2019-must} datasets (408 hours)\footnote{We also experimented with a smaller Europarl-ST dataset and to save space we report results in Appendix~\ref{sec:europarl}. Overall, they are similar to these on larger datasets.}. As expert model, we use the Transformer from Facebook's submission to WMT19~\citep{ng-etal-2019-facebook}, which is based on the Big Transformer architecture proposed by~\citep{vaswani2017attention}. Our sequence-to-sequence models for students are RNNs and Base Transformers. All models are based on the \texttt{fairseq} framework~\citep{ott-etal-2019-fairseq, wang-etal-2020-fairseq}, but use different settings of meta-parameters and preprocessing than the default models. More details on models, meta-parameters and training settings are given in the Appendix~\ref{app:details}.

Our training setups are summarized in Table~\ref{tab:overview}. We compare our trained student models with several baseline approaches: \enquote{Standard} denotes AST trained by teacher forcing on ground truth targets with a label smoothing~\citep{SzegedyVISW15} factor of 0.1. KD$^+$ \citep{liu19d_interspeech} denotes word-level knowledge distillation between the expert's and student's full output probability. IKD and IKD$^+$ denote imitation knowledge distillation, where student model is corrected by the empirical distribution of the optimal expert actions or the full expert policy~\citep{lin-etal-2020-autoregressive}, respectively. SynthIKD and SynthIKD$^+$ are our variants with synthetic transcripts. We used the same same exponential decay schedule ($\beta=\frac{1}{T})$ used by ~\cite{lin-etal-2020-autoregressive} as early experiments showed that this  performed best in our setup. 


\begin{table}[t!]
    \centering
    \resizebox{\linewidth}{!}{
    \begin{tabular}{llll}
    \toprule
    \textbf{Variant} & \textbf{Expert Input} & \textbf{Loss}\\
    \midrule
    Standard &  - &  CE\\
    \midrule
    KD$^+$~\cite{liu19d_interspeech}   & gold  & CE \\
    SynthKD$^+$ & synthetic  & CE \\
    \midrule
    IKD~\cite{lin-etal-2020-autoregressive} &  gold  & $\mathcal{L}_{\text{IKD}} $ \\
    IKD$^+$~\cite{lin-etal-2020-autoregressive} &  gold  & $\mathcal{L}_{\text{IKD}^+} $ \\
    \midrule
    SynthIKD (ours)  &  synthetic  & $\mathcal{L}_{\text{IKD}}$\\
    SynthIKD$^+$ (ours) &  synthetic  & $\mathcal{L}_{\text{IKD}^+}$\\
    \bottomrule
    \end{tabular}}
    \caption{Summary of training variants: \enquote{Standard} denotes AST trained via cross-entropy (CE) on ground truth targets with a label smoothing. KD$^+$ denotes word-level knowledge distillation between the expert's and student's full output probability. IKD and IKD$^+$ denote imitation knowledge distillation where student model is corrected by the optimal expert action or the full expert policy~\citep{lin-etal-2020-autoregressive}, respectively. SynthIKD and SynthIKD$^+$ are our variants with synthetic transcripts. Expert Input indicates whether the NMT expert is given the original transcripts from the dataset or synthetic transcripts created by ASR. All IKD methods use the exponential decay schedule for $\beta$ that \cite{lin-etal-2020-autoregressive} found to work best.}
    \label{tab:overview}
\end{table}

All AST models' encoders are initialized with the encoder of the corresponding ASR model, trained on the respective datasets with cross-entropy and the label-smoothing factor of 0.1. Because of the relatively small size of these datasets, our experiments should seen as proof-of-concept, showing that ASR models trained on a few hundred hours of audio provide synthetic transcripts of sufficient quality to enable imitation-based KD for AST. The standalone performance of our ASR models is listed in Table~\ref{tab:wers}. 

\begin{table}[t!]
 \small
	\centering
		\begin{tabular}{lrrrr}
			\toprule
			\multirow{2}{*}{\textbf{Model}}&  \multicolumn{2}{c}{\textbf{CoVoST2}} &  \multicolumn{2}{c}{\textbf{MuST-C}}\\
                & \textbf{dev} & \textbf{test} & \textbf{dev} & \textbf{test}\\
			\midrule
			RNN &   26.68 & 33.94 & 23.42 & 24.44\\
			Transformer & 20.93 & 26.60 & 21.10  & 20.68\\
			\bottomrule
		\end{tabular}
		\caption{WER$\downarrow$ results for ASR models pretrained on CoVoST2 and MuST-C. These models are used to create the synthetic transcripts for respective experiments. Standard development and test splits were used for CoVoST2. For MuST-C, we tested on \texttt{tst-COMMON}.}
	\label{tab:wers}
\end{table}


\subsection{Feasibility of Oracle Correction}

\begin{table*}[htb]\small
	\centering
		\centering
\resizebox{\textwidth}{!}{
		\begin{tabular}{cclllr}
			\toprule
			\textbf{Architecture} & \textbf{Hypotheses} & \textbf{\#} &\textbf{Decoding Setup}& \textbf{Source Transcripts} & \textbf{dev-BLEU$\uparrow$}\\
			\midrule
			  \multirow{4}{*}{RNN} & \multirow{2}{*}{full} & 1 &AST & - & 11.9\\
			                       &                      & 2 & ASR transcribes,  NMT expert translates & synthetic  & 21.8\\
			                     & \multirow{2}{*}{partial} & 3&AST starts, NMT expert completes & gold & 21.9\\
			 &&4&AST starts, NMT expert completes  & synthetic  & 15.6\\
			\midrule
			  \multirow{4}{*}{Transformer} & \multirow{2}{*}{full} 		  &5&AST & - &  16.7\\
			                              &                               &6&ASR transcribes,  NMT expert translates & synthetic  & 25.4\\
			                             &\multirow{2}{*}{partial} &7&AST starts, NMT expert completes  & gold & 25.4\\
			                           &                         &8&AST starts, NMT expert completes  & synthetic & 19.9\\
   \bottomrule
		\end{tabular}}
		    \caption[Results of NMT expert completions]{Feasibility experiment: BLEU score on CoVoST2 development set of NMT expert's completion of AST model full or partial hypotheses with greedy decoding; \textit{gold} denotes the usage of the dataset's source language transcripts as NMT inputs and \textit{synthetic} denotes synthetic transcripts created by the respective ASR model.}
\label{tab:bleu_devs}
		\end{table*}
  
The idea of using synthetic transcripts in place of gold transcripts has merit only if the NMT oracle's translations have higher quality than the translations the AST model generates. Therefore, we first verify if the NMT oracle is capable of completing an AST models' partial hypotheses $y_{<t}$ while improving quality at the same time. 

We follow~\citet{lin-etal-2020-autoregressive} and let the AST models trained with label-smoothed CE on ground truth targets translate the audio input with greedy decoding up to a randomly chosen time step. 
Then, we feed the NMT expert the gold transcript as input and the partial translation as prefix, and let the oracle finish the translation with greedy decoding. 

As Table~\ref{tab:wers} shows, the out-of-the-box ASR performance is relatively low (high WER), so errors in synthetic transcripts will be propagated through the NMT oracle. The question is whether the expert's continuation can be of higher quality than the student's own predictions despite the partially incorrect synthetic transcripts. 
In Table~\ref{tab:bleu_devs}, lines 1 and 2 (or, 5 and 6) set the lower (end-to-end) and upper (cascade) bounds on the performance. We see that the NMT expert is able to complete the student hypotheses successfully (lines 3, 4 and 7, 8), bringing gains in both gold and synthetic setups, and reaching the upper bound (lines 3 vs. 2 and 7 vs. 6) for gold ones. Although the mistakes in the synthetic transcripts do result in lower BLEU scores (lines 4 and 8) they still improve over the AST student complete translations (lines 1 and 5).

\subsection{Main Results}\label{sec:main_results}

 Table~\ref{tab:imitkd_results} shows the main results of applying Algorithm~\ref{alg:pseudo} for training an AST  student with imitation-based knowledge distillation on CoVoST2 and MuST-C.

\begin{table}[htb]
  \centering
  \resizebox{\linewidth}{!}{
	\begin{tabular}{lllrr|rr}
		\toprule
				\multirow{2}{*}{\textbf{Achitecture}} & \multicolumn{2}{c}{\multirow{2}{*}{\textbf{Models}}} &  \multicolumn{2}{c|}{\textbf{CoVoST2}} & \multicolumn{2}{c}{\textbf{MuST-C}}\\ 
				& & & \textbf{dev} & \textbf{test} & \textbf{dev} & \textbf{test}\\
		\midrule
		 \multirow{5}{*}{RNN} & \multirow{3}{*}{\rotatebox[origin=c]{90}{baseline}}&Standard & 13.6 & 10.0 &  14.6 & 14.1\\
		&&KD$^+$  & \textbf{14.6} & \textbf{11.1} &\textbf{17.9} & \textbf{17.2}\\
		&&IKD$^+$  & 13.1 & 10.1& 15.7 & 14.9\\
		&\multirow{2}{*}{\rotatebox[origin=c]{90}{ours}}&SynthKD$^+$   & 14.1 & 10.6 & 16.9 & 15.9\\
		&&SynthIKD$^+$ &  12.8  & 9.7&  16.3 & 15.1\\
		\midrule
		\multirow{5}{*}{Transformer} & \multirow{2}{*}{\rotatebox[origin=c]{90}{baseline}}&
		Standard  & 18.4 & 14.2 &  19.5 & 19.4\\
		&&KD$^+$ &  21.3 & 17.7 &  17.7 & 22.2 \\
        &&IKD$^+$  &  \textbf{21.8} & 18.4& 23.2 & 23.3 \\
		&\multirow{2}{*}{\rotatebox[origin=c]{90}{ours}}&SynthKD$^+$  &  21.7 & 18.0 &   22.5 & 22.6 \\
		&&SynthIKD$^+$  &  \textbf{21.8} & \textbf{18.5}&  \textbf{23.5} & \textbf{23.5}\\
		\bottomrule
	\end{tabular}}
 	\caption{Main results: RNN and Transformer student models trained on expert inputs and loss variants of Table \ref{tab:overview}, using Dagger for IL.
  We used the \texttt{tst-COMMON} as the test set for \mbox{MuST-C}. (Synth)IKD is not included since its performance is worse than (Synth)KD$^+$. Transformers trained with IL outperform all baselines, while pure KD is the best for generally lower-quality RNN-based models. Synthetic transcripts do not harm performance for Transformer student models.
  }\label{tab:imitkd_results}
 \end{table}

\paragraph{Dagger} First we present results for the Dagger algorithm. In Table~\ref{tab:imitkd_results}, for both CoVoST2 and \mbox{MuST-C} models, Dagger with the Transformer architecture outperforms all baselines\footnote{$p$-value $< 0.005$ using the paired approximate randomization test~\cite{riezler-maxwell-2005-pitfalls}}, and matching full teacher distributions (the `+'-versions of losses) gives consistent gains. Imitation-based knowledge distillation with RNNs, on the other hand, fails to improve BLEU scores over baselines, most likely due to their overall lower translation quality. This leads to the student hypotheses that are too far from the reference so that the expert's one-step corrections are not able to correct them. 

The results show that Transformers and RNNs with synthetic transcripts show statistically insignificant differences in performance to the ones that are using gold transcripts. This is notable since the partially synthetic transcripts provided to the NMT oracle are often incorrect, yet do not result in a noticeable effect on the final student performance if used in the IL framework. 

\paragraph{AggreVaTe} Finally, we evaluate the performance of AggreVaTe both with gold and synthetic transcripts. During training we targeted and  evaluated with the non-decomposable BLEU metric (i.e. training with sentence-BLEU and evaluating with corpus-BLEU) as well as with the decomposable TER metric (Table~\ref{tab:AggreVaTe}). Following~\citet{DBLP:journals/corr/abs-2109-04114} we warm-started AggreVaTe with differently trained standard or Dagger models, and trained with AggreVaTe objectives for up to 50 epochs with early stopping on respective development sets. 

Surprisingly, we found that AggreVaTe does not bring additional benefits on top of Dagger despite the promise for a better matching between training and inference objectives. Also there is no significant difference between the results with the TER rewards objective and sentence-BLEU rewards on both CoVoST2 and MuST-C. We explain these results by the sufficiency of one-step corrections to correct a ``derailed'' student, with little benefit of continuing demonstration till the end of translation. 
The fact that Dagger turns out to reap all of the benefits from training with IL is good news in general, since running beam search during training (to get AggreVaTe's full continuations) is more expensive than greedily selecting one action (as does Dagger). 


\begin{table*}[!ht]
	\centering
\small
	\begin{tabular}{lllrrrr|rrrr}
	\toprule
		\multirow{3}{*}{\textbf{IL Algorithm}} & \multirow{3}{*}{\textbf{Model}} &  \multirow{3}{*}{\textbf{Data}} & \multicolumn{4}{c|}{\bf CoVoST2} &  \multicolumn{4}{c}{\bf MuST-C} \\
  &&& \multicolumn{2}{c}{\textbf{BLEU$\uparrow$}} & \multicolumn{2}{c|}{\textbf{TER$\downarrow$}} & \multicolumn{2}{c}{\textbf{BLEU$\uparrow$}} & \multicolumn{2}{c}{\textbf{TER$\downarrow$}}\\
	               &&&  \textbf{dev} & \textbf{test} & \textbf{dev} & \textbf{test} & \textbf{dev} & \textbf{test} & \textbf{dev} & \textbf{test}\\
   \midrule
 	\multirow{3}{*}{Dagger} & Standard & gold   & 18.4 & 14.2  & 69.1 & 77.1&  19.5 & 19.4   &  70.8 & 69.4\\
     	                        & IKD$^+$  & gold   & 21.8 & 18.4  & 63.7 & 70.0&  23.2 & 23.3   & 67.4 & 65.6\\
		                        & SynthIKD$^+$ & synth & 21.8 & \textbf{18.5} & 63.6 & 69.8& \textbf{23.5} & 23.5 &  67.2 & 65.6 \\

   \midrule
   		
           &\multirow{2}{*}{\textbf{Warm-start Model}} & \multirow{2}{*}{\textbf{Data}} & \multicolumn{2}{c}{\textbf{BLEU$\uparrow$}} & \multicolumn{2}{c|}{\textbf{TER$\downarrow$}} & \multicolumn{2}{c}{\textbf{BLEU$\uparrow$}} & \multicolumn{2}{c}{\textbf{TER$\downarrow$}}\\
	&&&  \textbf{dev} & \textbf{test} & \textbf{dev} & \textbf{test} & \textbf{dev} & \textbf{test} & \textbf{dev} & \textbf{test}\\
  \midrule
			\multirow{10}{*}{AggreVaTe}&\multicolumn{10}{c}{sentence-BLEU reward-to-go}\\
     		 & Standard & gold & 18.7 & 14.6 & 68.2 & 76.0& 19.9 & 19.9 & 70.2 & 68.1\\
			& Standard & synth &  18.7 & 14.6 & 68.2 &75.9&  20.0 & 19.7 & 70.1 & 68.7\\
          & IKD$^+$ & gold & \textbf{22.1} & \textbf{18.5} & \textbf{63.1} & 69.6& \textbf{23.5} & 23.4 & 67.4 & 65.7\\
          & SynthIKD$^+$ & synth  & \textbf{22.1} & \textbf{18.5} & \textbf{63.1} &  69.7&  \textbf{23.5} & \textbf{23.6} & \textbf{67.0} & 65.6\\
			&\multicolumn{10}{c}{TER reward-to-go}\\
     		&Standard & gold & 18.7 & 14.7 & 67.8 & 75.4& 20.0 & 19.9 & 70.0 & 68.5\\
			&Standard & synth &  18.7 & 14.6 & 67.9 & 75.6&  19.9 & 19.6  & 69.8 & 68.4\\
          & IKD$^+$  & gold & 22.0 & \textbf{18.5} & \textbf{63.1} & \textbf{69.4}& 23.3 &  23.4 & 67.3 & 65.5\\
          & SynthIKD$^+$ & synth  & \textbf{22.1} & \textbf{18.5} & \textbf{63.1} &  69.6& \textbf{23.5}  & \textbf{23.6} & \textbf{67.0} & \textbf{65.3}\\

			\bottomrule
		\end{tabular}
	\caption{Comparison of Dagger with warm-started AggreVaTe with a maximum of 50 epochs on CoVoST2 and MuST-C.}\label{tab:AggreVaTe} 
\end{table*}

\subsection{Quality of Synthetic Transcripts}

\begin{table}[!hbt]
	\centering
\small
	\begin{tabular}{lrrrr}
		\toprule
				\multirow{2}{*}{\textbf{Training}} &  \multicolumn{2}{c}{\textbf{CoVoST2}} &  \multicolumn{2}{c}{\textbf{MuST-C}}\\ 
				& \textbf{dev} & \textbf{test} & \textbf{dev} & \textbf{test}\\
		\midrule
		\multicolumn{5}{c}{\textbf{training on translated gold transcripts}}\\

		 Standard & 18.1 & 14.9&  20.0 & 20.0\\
	 KD$^+$  &  21.3 & 17.6&   23.4 & 23.1  \\
		IKD$^+$  & 22.6 & 18.6& 23.5 & 23.7\\
		\midrule
		\multicolumn{5}{c}{\textbf{training on translated synthetic transcripts}}\\
		
		Standard   & 17.8 & 14.2&  19.2 & 19.2\\
		KD$^+$   &  20.2 & 16.5&   22.1 & 22.5    \\
		IKD$^+$  & 21.0 & 17.4& 23.0 & 23.1  \\
		\bottomrule
	\end{tabular}
	\hfill
	\caption{BLEU scores of Transformer models trained on the training set with original references replaced by translations of gold and synthetic transcripts in comparison to using the original training set (lower part of Table~\ref{tab:imitkd_results}). 
 } 
 	\label{tab:sequence_level_kd}
\end{table}

In this section, we investigate explanations for the high performance of Dagger on synthetic transcripts: The first hypothesis is that synthetic transcripts are already ``good enough'' and per-step IL corrections add nothing on top. Second, the gains could be due to the known NMT ``auto-correcting'' ability and due to general robustness to the quality of the source (cf. the success of back-translation in NMT), and all benefits could be reached with KD alone.
To test both hypotheses, we create new training datasets where we replace references with translated gold or synthetic transcripts by the same NMT expert with beam size 5.
Evaluating on the unmodified references, we trained Transformer-based baselines and the IL model from~\citet{lin-etal-2020-autoregressive} on these two new corpora. 

As Table~\ref{tab:sequence_level_kd} shows, 
Transformer KD$^{+}$ trained on translated gold transcripts outperforms its counterparts trained on translated synthetic transcripts, confirming errors in the synthetic transcripts. This refutes the first hypothesis.

Regarding the second hypothesis, we compare the KD$^+$ to IKD$^+$ from the synthetic translated part in Table~\ref{tab:sequence_level_kd}. Were ``auto-correction'' sufficient we would see similar performance in both lines. This rejects the second hypothesis and suggests that IL adds value on top of general NMT robustness to inputs.

\subsection{Qualitative Analysis}

Here, we perform a human evaluation of successful IL corrections, aiming at an explanation of the performance of Dagger on synthetic transcripts. 

\begin{figure}[!hbt]
\centering
	\includegraphics[scale=0.12]{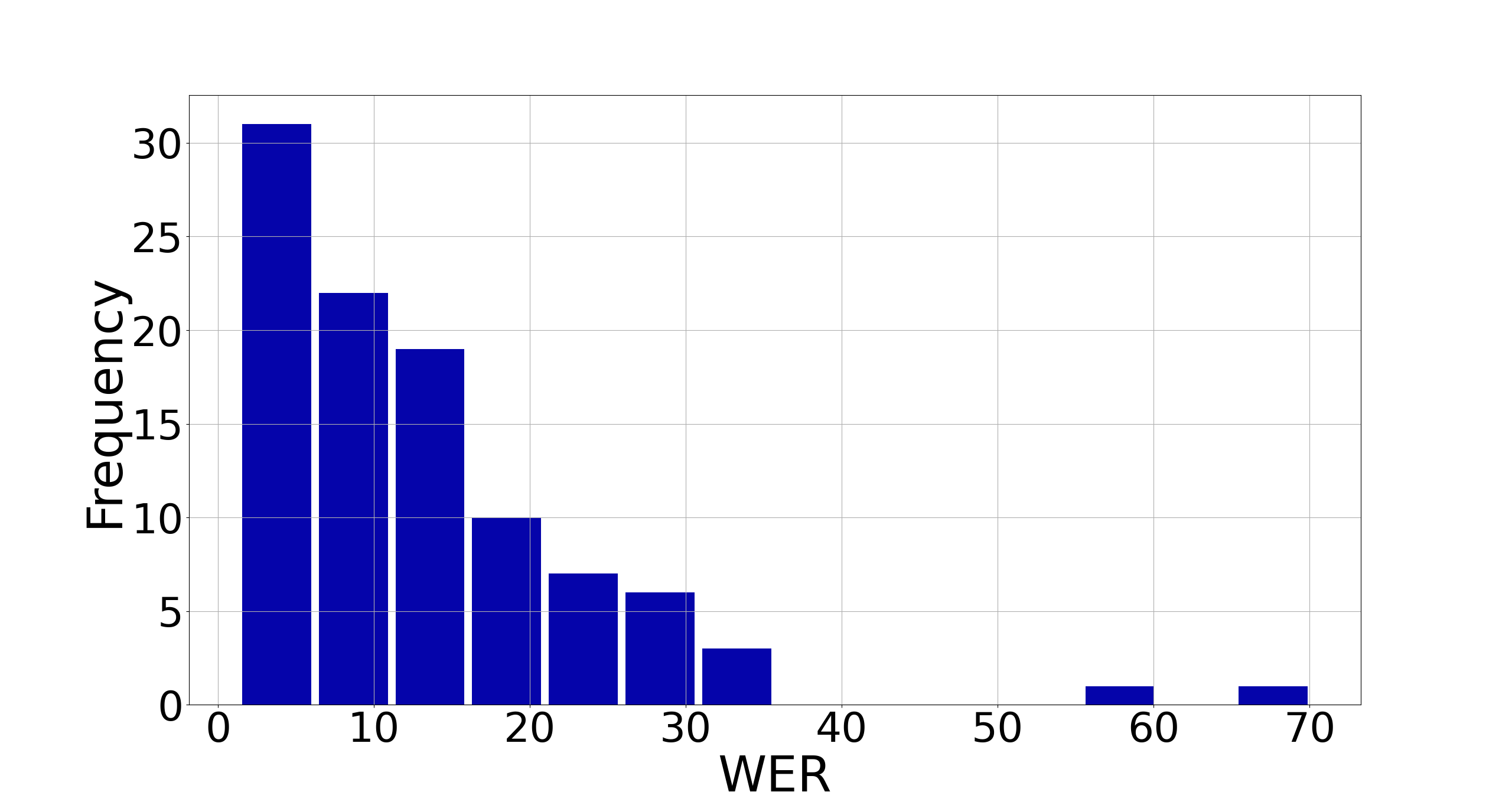}
	\caption{Histogram of sentence-wise WER of ASR Transformer on 100 samples from CoVoST2.}\label{fig:histo}
\end{figure}

We randomly sample 100 examples from the CoVoST2 training set on which the ASR Transformer has a non-zero sentence-wise word error rate, and compare the NMT expert's probability distributions over time for the given synthetic transcripts. From the WER histogram in Figure~\ref{fig:histo} we see that most of the sentences have a single-digit number of errors.

As WER cannot be used to differentiate between small but inconsequential (to the understanding of the sentence) errors and mistakes that change the meaning of the sentence, we further compare the generated transcript to the gold transcript \textit{and} look at the top-8 output probabilities of the expert at each time step for each sample to classify each error in the synthetic transcripts.
We further feed the sampled sentences to the NMT expert and find that in 36 out of 100 samples (all but the last two lines in  Table~\ref{tab:errors}), the expert is able to generate output probability distributions that favor the correct target token despite errors in the transcript. 
Although the expert can put large probability mass on the correct target token, whether it does so depends on the error type in the generated transcript. The expert is often able to deal with surface form errors, such as different spellings, punctuation errors and different word choice (17 occurrences).
When the synthetic transcripts contain critical errors, e.g. partially hallucinated transcript, the expert is still able to produce the correct translation if the missing or wrong information can be still inferred from the prefix (32 occurrences).
\begin{table}[t!]
	\centering
	\small
\begin{tabular}{p{6.5cm} r}
		\toprule
		\textbf{Error Type} & \textbf{Freq}\\
		\midrule
      omitted tokens  & 2\\
		surface form error & 17 \\
       contentual error, correct target in top-1 & 5 \\
       contentual error, correct target in top-8 & 12\\
		critical error, expert predicts correctly due to prefix & 32 \\
            critical error, expert does not predict correctly & 32 \\
		\bottomrule
\end{tabular}
\caption{Error types in the synthetic transcripts created by the ASR model.}
\label{tab:errors}
\end{table}

\begin{figure*}[t!]
    \begin{minipage}{.48\linewidth}
	\centering
	\includegraphics[scale=0.15]{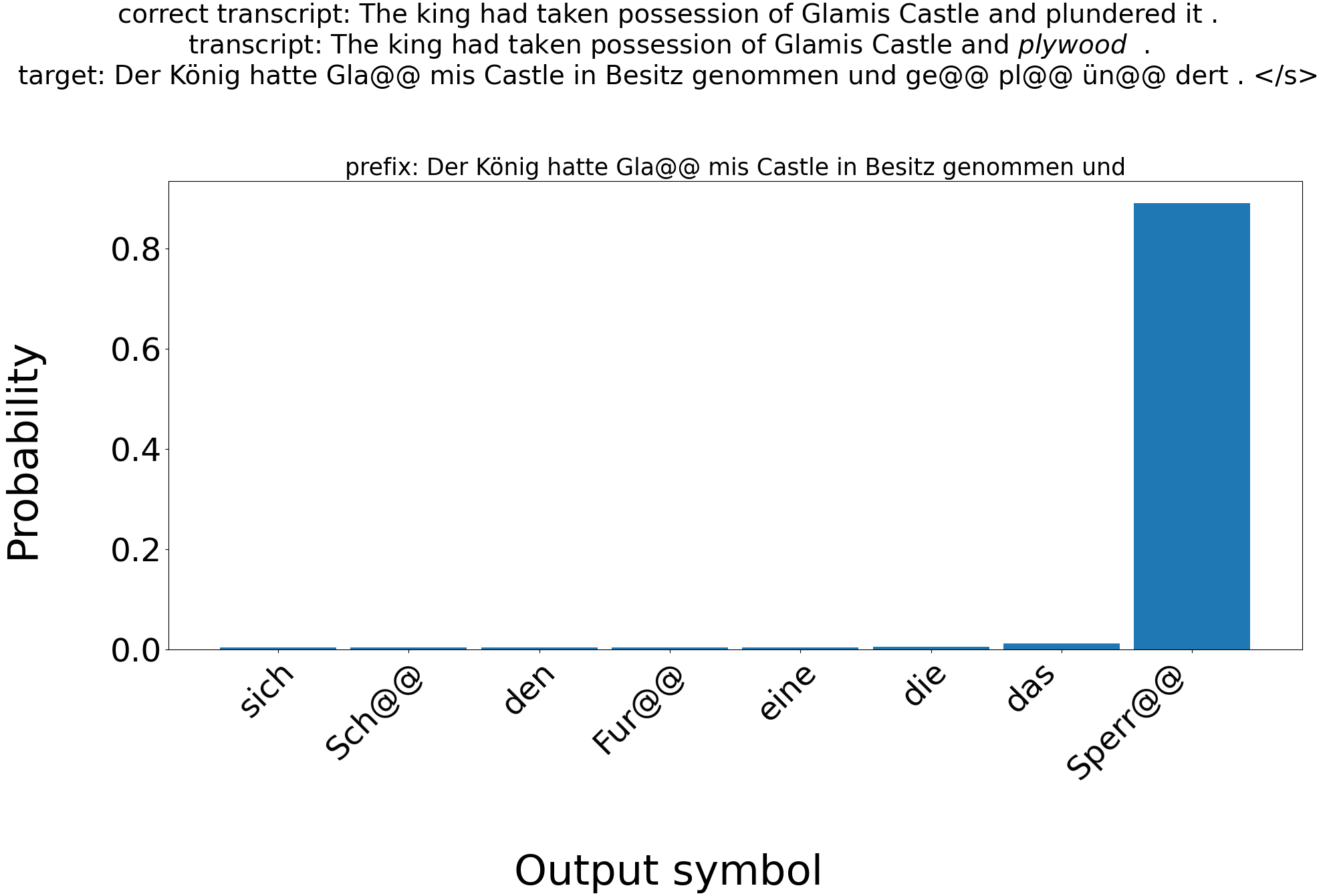}
	\subcaption{with $y_{<t}$ = \enquote{Der König hatte Glamis Castle in Besitz genommen und }}
	\label{fig:king_sperr}
	\end{minipage}
	\hfill
    \begin{minipage}{.48\linewidth}
	\centering
	\includegraphics[scale=0.15]{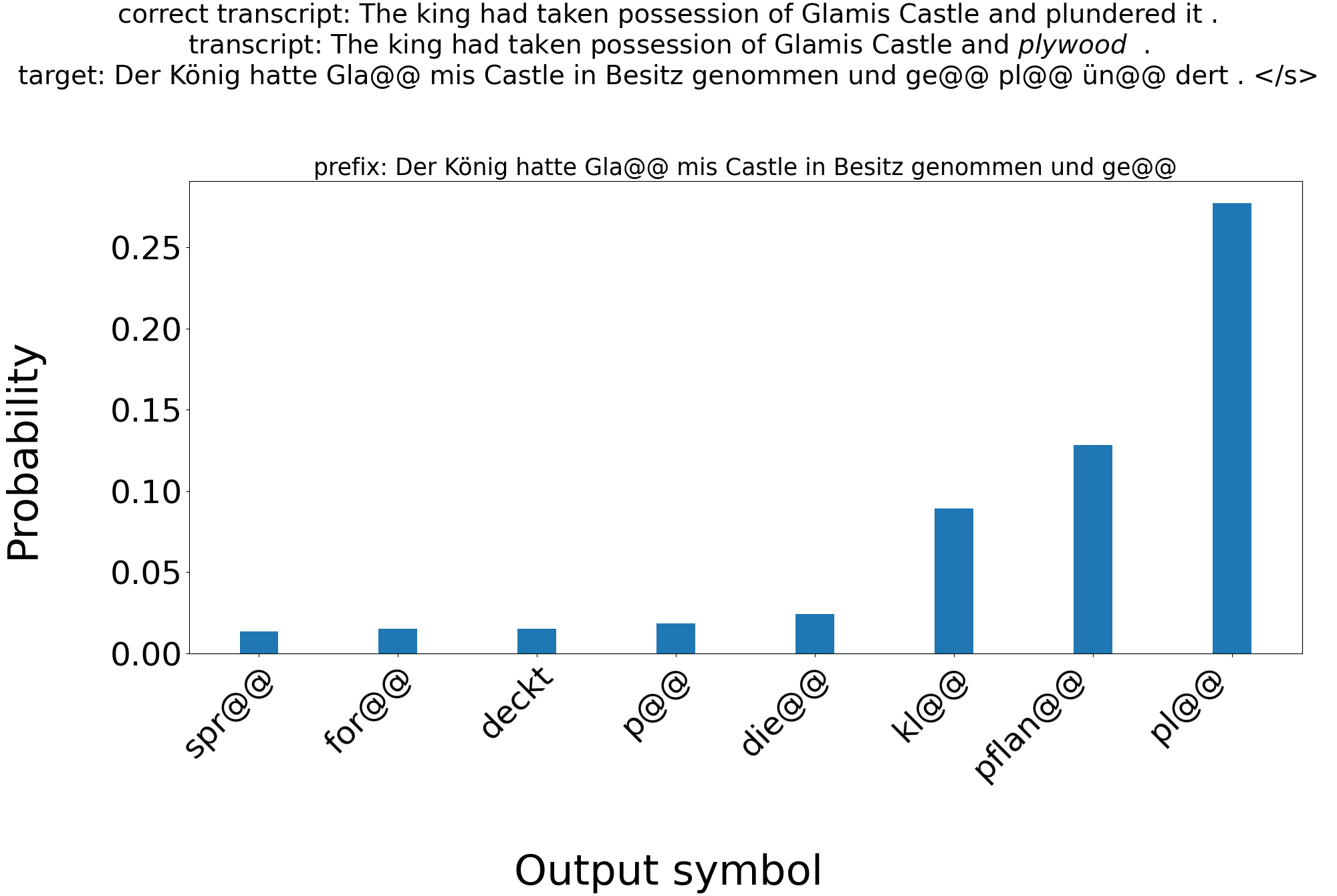}
	\subcaption{with $y_{<t}$ = \enquote{Der König hatte Glamis Castle in Besitz genommen und ge}}
	\label{fig:king_ge}
	\end{minipage}
	\caption{NMT expert top-8 output probabilities when translating the incorrect synthetic transcript \enquote{The king had taken possession of Glamis Castle and plywood it.}}
	\label{fig:king}
\end{figure*}

Next, we verify that the decoder language modeling capability is what primarily drives the correction process. We do this by feeding parts of reference translations as prefix conditioned on erroneous synthetic transcripts. 
Consider the transcript \enquote{The king had taken possession of Glamis Castle and plywood.} generated by the ASR model. Its gold transcript reads \enquote{plundered it} instead of \enquote{plywood}.
In Figure~\ref{fig:king} we illustrate output probabilities that the expert generates in the last time-steps.
\begin{figure*}[t!]
    \begin{minipage}{.48\linewidth}
		\centering
		\includegraphics[scale=0.15]{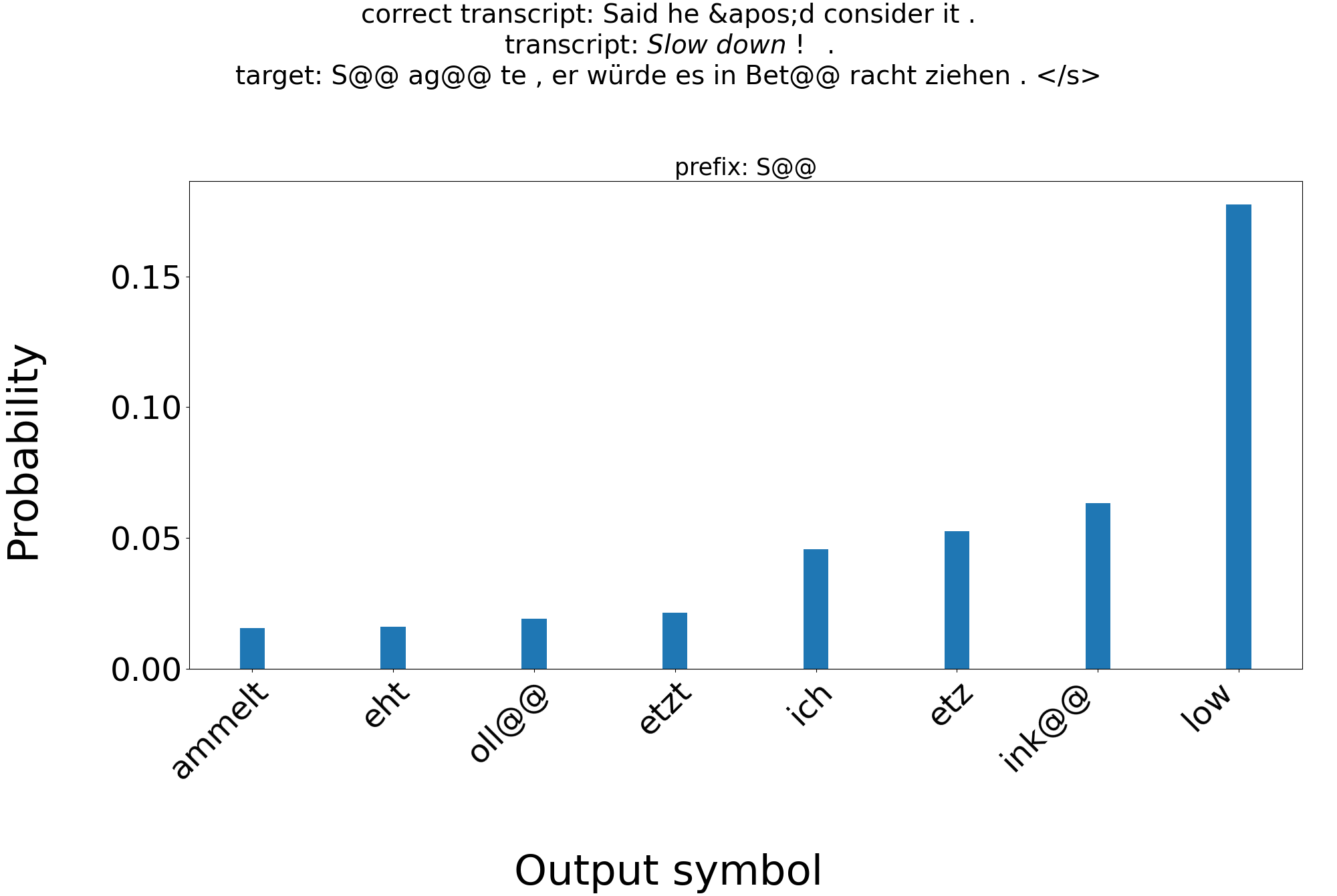}
		\subcaption{with $y_{<t}$ = \enquote{S}}
		\label{fig:slow_down_s}
	\end{minipage}\hfill
	 \begin{minipage}{.48\linewidth}
		\centering
		\includegraphics[scale=0.15]{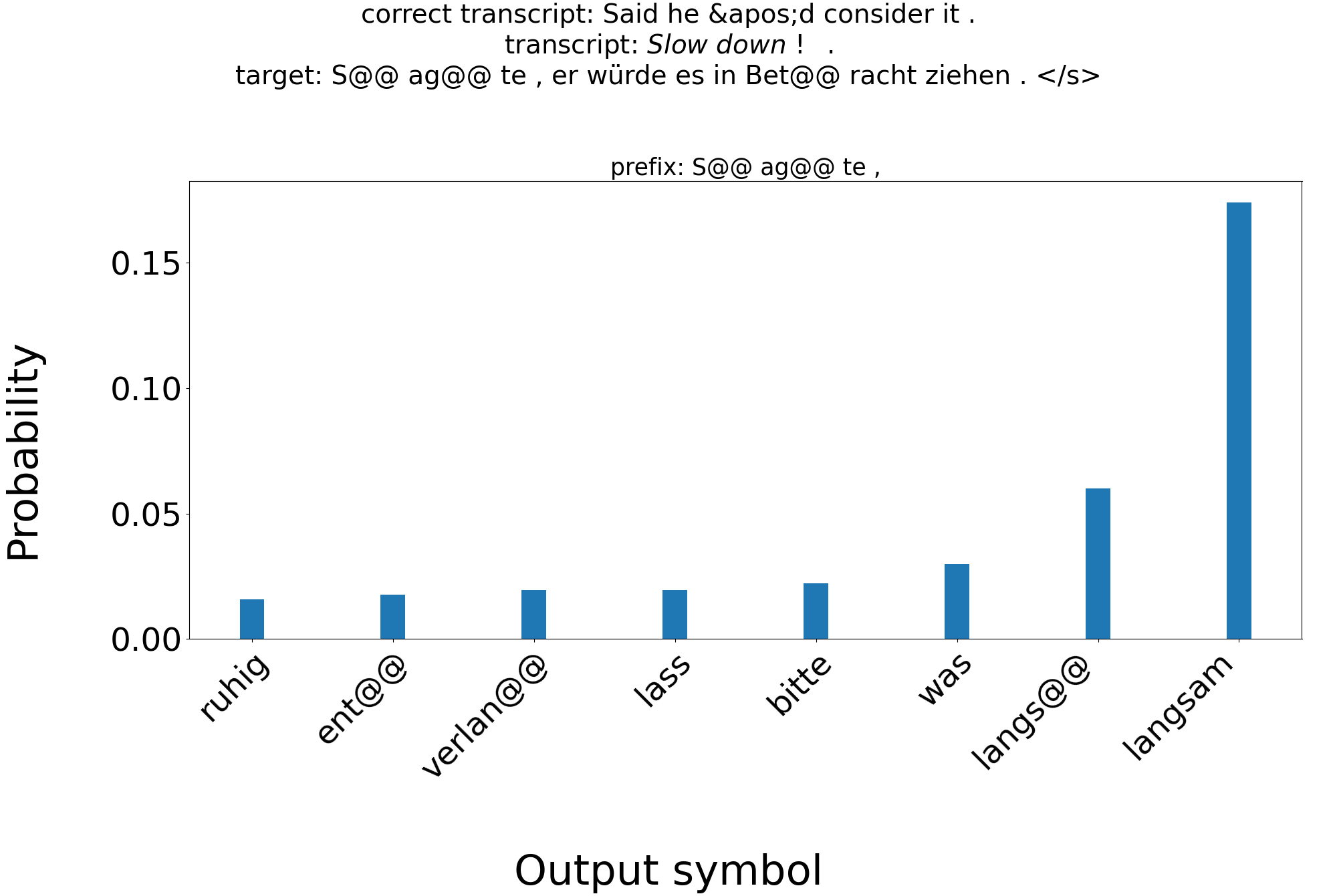}
		\subcaption{with $y_{<t}$ = \enquote{Sagte , }}
		\label{fig:slow_down_sagte}
	\end{minipage}
	\caption{NMT expert top-8 output probabilities when translating the incorrect synthetic transcript \enquote{Slow down!}}
	\label{fig:slow_down}
\end{figure*}
Assume as in Figure~\ref{fig:king_sperr} that the expert has been given the prefix \enquote{Der König hatte Glamis Castle in Besitz genommen und}. According to the output probabilities, the next output symbol is the subword unit \enquote{Sperr} and would not be a proper correction. At the next timestep, however, the last symbol in the prefix is the subword unit \enquote{ge} and,
as Figure~\ref{fig:king_ge} shows, the expert, being driven by its decoder language modeling capability, puts highest probabilities on subword units that are most likely to produce a fluent output (the correct one \enquote{pl@@}, and less probable \enquote{pflan@@} and \enquote{kl@@} 
rather then paying attention to the (wrong) information in the synthetic transcripts.


Similar situations can be observed in samples with entirely wrong synthetic transcripts.
In Figure~\ref{fig:slow_down}, the expert has received the synthetic transcript \enquote{Slow down!} as input, which shares no meaning with the gold transcript \enquote{Said he'd consider it.}
As shown in Figure~\ref{fig:slow_down_s}, the expert assigns the highest probability to \enquote{@@low} if it is given the prefix \enquote{S} (as the expert has a shared vocabulary, it can complete the output this way), which turns the partial translation into an exact copy of the transcript.
Again, the top-8 predictions do not share similar meaning with the transcript.
After, in Figure~\ref{fig:slow_down_sagte}, the expert has received the prefix \enquote{Sagte,}, it still attempts to complete $y_{<t}$ by generating output symbols that would turn $y$ into a valid translation of this wrong transcript (\enquote{langsam} (slow), \enquote{ruhig} (quiet), \enquote{langs@@})) with the rest of options being mostly driven by language modeling rather then reproducing source semantics (\enquote{ent@@}, \enquote{verlan@@}).

Overall, 
with the SynthIKD$^+$ training, the expert induces smoothed 
output distributions and fluency on the student
more than it enforces 
the student to predict one-hot labels produced by the expert as is done by sequence-level KD.



\section{Conclusion}

We showed that a pretrained NMT model can successfully be used as an oracle for an AST student, without requiring gold source language transcripts as in previous approaches to imitation learning for AST. This widens the applicability of imitation learning approaches to datasets that do not contain manual transcripts, or to the use of pre-trained ASR models for which training transcripts are not available. Our qualitative analysis suggests an explanation of the fact that the NMT oracle is robust against mismatches between manual and synthetic transcripts by its large language model capabilities that allow it to continue the prefix solely based on its learned contextual knowledge.


\section{Limitations}

There are several limitations of this study. First, it is done on one language pair although we believe this should not qualitatively change the results. Second, only one set of standard model sizes was evaluated for AST student and  NMT expert; we expect it be in line with reported findings for NMT~\cite{ghorbani}. Finally, while alluding to the potential of using large pre-trained ASR models instead of manual transcripts for IL-based AST, our current work must be seen as a proof-of-concept experiment where we train ASR models on a few hundred hours of audio, and discard the manual transcripts in IL training, showing the feasibility of our idea. 


\section*{Acknowledgements}

The authors acknowledge support by the state of Baden-Württemberg through bwHPC and the German Research Foundation (DFG) through grant INST 35/1597-1 FUGG.

\bibliography{anthology,custom}
\bibliographystyle{acl_natbib}
\clearpage
\appendix

\counterwithin{figure}{section}
\counterwithin{table}{section}
\newpage
\section{Models, Meta-parameters, and Training Settings}\label{app:details}

\begin{table}[t!]
	\centering
	\begin{tabular}{lrr}
		\toprule
				\multirow{2}{*}{\textbf{Model}} &  \multicolumn{2}{c}{\textbf{BLEU$\uparrow$}} \\
				 & \textbf{dev} & \textbf{test} \\
		\midrule
  \multicolumn{3}{c}{\textbf{original dataset}}\\
  
		Standard & 13.8& 14.4  \\
		KD$^+$  & 17.4 & 17.8\\
		SynthKD$^+$  & 17.5 & 18.0 \\
		IKD$^+$  & 17.0 & 17.1\\
		  SynthIKD$^+$ &  17.0 & 17.0\\
		
		\multicolumn{3}{c}{\textbf{translated gold training set}}\\
		
		Standard & 15.3  & 15.3  \\
		KD$^+$    &\textbf{18.2} &\textbf{18.4} \\
		IKD & 16.8 & 17.0\\
		IKD+ & 17.1  & 17.5  \\
		
		\multicolumn{3}{c}{\textbf{synthetic translated training set}}\\
		
		Standard  &  14.7 & 15.3  \\
		KD$^+$   &17.0 & 16.8 \\
		IKD & 16.1 & 16.0  \\
		IKD+ & 16.3 & 16.6  \\
		\bottomrule
	\end{tabular}
	\caption{Results on Europarl-ST}
	\label{tab:europarl_results}
\end{table}
We use the speech-to-text module of the \texttt{fairseq} framework~\citep{ott-etal-2019-fairseq, wang-etal-2020-fairseq} for all experiments and train both RNNs with convolutional layers for time dimension reduction as in~\citet{DBLP:conf/icassp/BerardBKP18} and small Transformers as in~\citet{wang-etal-2020-fairseq}, which consist of a convolutional subsampler of two convolutional blocks, followed by 12 encoder layers and 6 decoder layers. The dimension of the self-attention layer is 256 and the number of attention heads is set to 4.
For the NMT oracle, we use the trained Transformer model from the Facebook's submission to WMT19~\citep{ng-etal-2019-facebook} \footnote{As the WMT19 submission consists of an ensemble of models, we use the \texttt{model1.pt} for our experiments.}, which is based on the big Transformer~\cite{vaswani2017attention} which has 6 encoder and decoder layers, 16 attention heads and the dimension of 1024, with a larger feed-forward layer size of 8192. This NMT oracle had been trained on all available WMT19 shared task en-de training data and on  back-translated english and german  portions of the  News crawl dataset. 

For all models we use Adam~\citep{Kingma2015AdamAM} with gradient clipping at norm 10 and stop training if the development set loss has not improved for 10 epochs. For RNN architectures, we return the best model on the development set and for Transformers, 
we create each model by averaging over the last 10 checkpoints.
For inference, a beam size of 5 was used and we report case-sensitive detokenized BLEU~\citep{papineni-etal-2002-bleu} computed with sacreBLEU~\citep{post-2018-call}. We tested for statistical significance with the paired approximate randomization test~\cite{riezler-maxwell-2005-pitfalls}.

For all experiments, we preprocess the datasets as follows:
We extract log mel-scale filterbanks with a povey window, 80 bins, a pre-emphasis filter of 0.97, a frame length of 25 ms and a frame shift of 10 ms.
We discard samples with less than five or more than 3000 frames and subtract the mean of the waveform from each frame and zero-pad the FFT input. 
For the text data, we normalize punctuation, remove non-printable characters, use the Moses tokenizer~\citep{koehn-etal-2007-moses} for tokenization and  segment the text data into subword units with byte-pair encoding~\citep{sennrich-etal-2016-neural}.
We used a random seed of 1 for all experiments.     

We list the final used and best performing hyperparameters in Table~\ref{tab:hyperparams}. Parameters that do not differ between the training methods are not repeated in the table. 
We determine the batch size by defining a maximum number of input frames in the batch. 

\begin{table*}[hbt]
	\centering
	\begin{tabular}{lrrr}
		\toprule
		\textbf{Model} & \textbf{Hyperparameter} & \textbf{CoVoST2} & \textbf{MuST-C}\\
		\midrule 
		\multicolumn{4}{l}{\textbf{RNN}}\\
		standard & learning rate & 1e-3 & 1e-3\\
  		& max-tokens & 60000 & 40000\\
		& scheduler & fixed & fixed\\
		& warmup-updates & 20000 & 20000\\
		& encoder freezing updates & 10000 & 10000\\
		& dropout & 0.2 & 0.2 \\
  
		KD$^+$ & learning rate & 1e-3 & 2e-3 \\
		& max-tokens & 50000 & 30000\\
		& warmup-updates & 25000 & 20000\\
		& max-update & 250000 & 250000\\
		& encoder-freezing updates & 20000 & 10000\\
		& scheduler & inverse square root & inverse square root\\
				
    \midrule
		\multicolumn{4}{l}{\textbf{Transformer}}\\

		 \textbf{ASR}  & learning rate & 2e-3 & 1e-3\\
		& max-tokens & 50000 & 40000\\
		& max-update & 60000 & 100000\\
		& scheduler & inverse square root & inverse square root \\
		& warmup-updates & 10000 & 10000\\
		& dropout & 0.15 & 0.1 \\
 
    \textbf{AST}\\
    
		standard & learning rate & 2e-3 & 2e-3 \\
		& max-update & 30000 & 100000\\
		& encoder-freezing updates & 1000 & -\\
 
		KD$^+$ & max-tokens & 50000 & 20000\\
        \bottomrule
	\end{tabular}
	\caption[List of hyperparameters]{list of hyperparameters that are dependent on model and dataset; we list only parameters which differ from the previous model's}
	\label{tab:hyperparams}
\end{table*}

\section{Europarl-ST}\label{sec:europarl}
We performed additional experiments on the Europarl-ST dataset~\citep{9054626} that provides 83 hours of speech training data.
We train RNNs with a learning rate of 0.002 and a max-tokens size of 40,000 for a total of 80,000 updates. All other hyper-parameters are the same as listed for MuST-C in Table~\ref{tab:hyperparams}.
We only trained RNNs on the Europarl-ST dataset due to the small amount of available training data. 
We present the results in Table~\ref{tab:europarl_results}.

Both improvements over standard training and by training on both the gold-translated and synthetic-translated translated training data correspond with the results presented in the main body of this work.
Hence, the results presented here hold for relatively small datasets, too.

\section{Additional Example of NMT Expert Correction}\label{app:examples}

\begin{figure}[htb]
	\centering
	\includegraphics[scale=0.15]{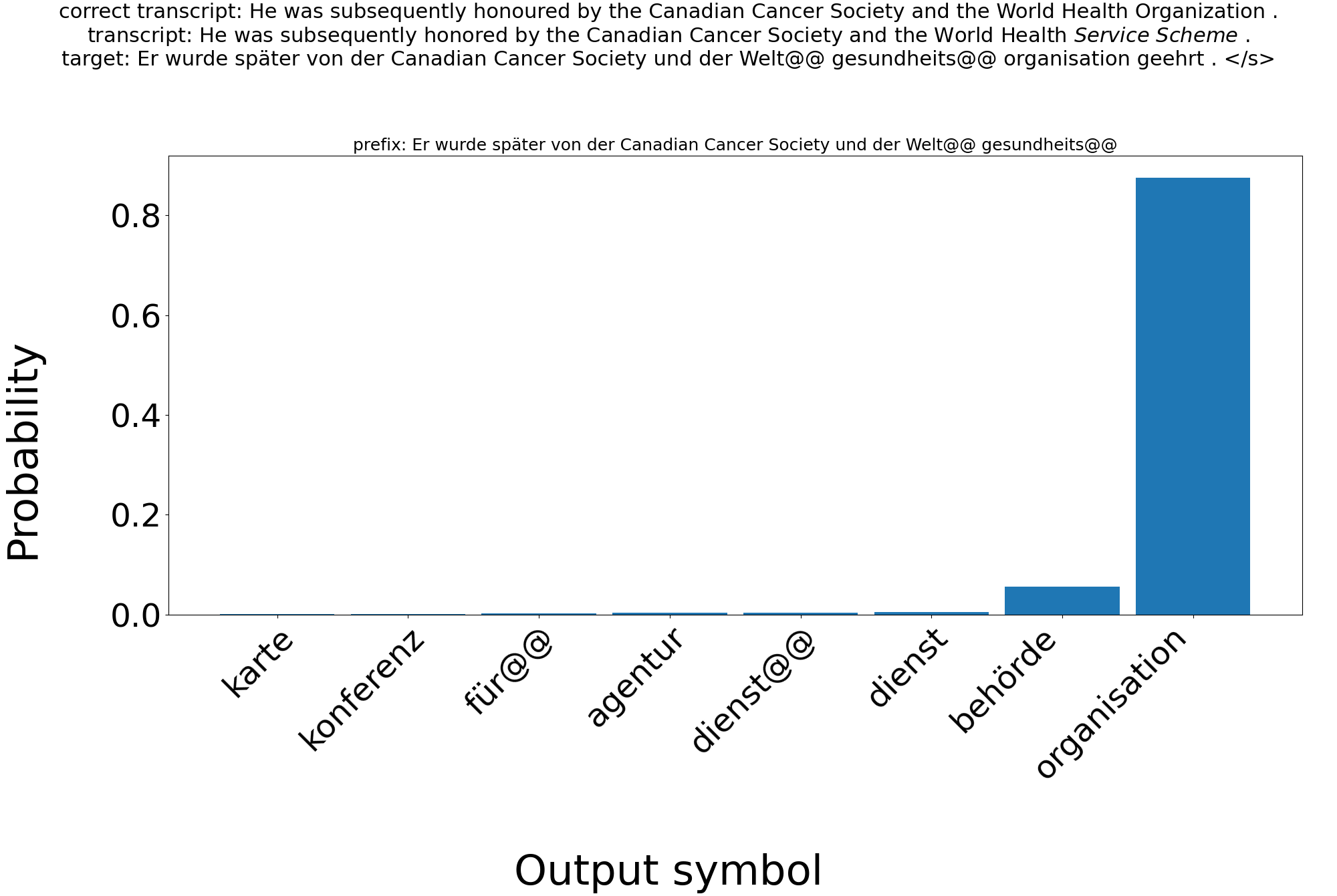}
	\caption{NMT expert top-8 output probabilities with $y_{<t}$ = \enquote{ Er wurde später von der Canadian Cancer Society und der Weltgesundheits}.}
	\label{fig:who}
\end{figure}
Here we give another example of the NMT expert predicting the correct output token despite receiving a transcript with incomplete or false information.
 
Figure~\ref{fig:who} shows the expert's output probabilities in response to receiving factually false information in the transcript.
The ASR model transcribed \enquote{World Health Organization} as \enquote{World Health Service Scheme}, yet the  expert produces a probability distribution that is skewed in favor of the correct proper name due to its learned context knowledge.
Note that the probability of generating the correct output token \enquote{organisation} (organization) is above 0.8.

\end{document}